\PassOptionsToPackage{table,dvipsnames}{xcolor}
\PassOptionsToPackage{hyphens}{url}
\documentclass[fleqn,10pt]{wlscirep}
\usepackage[utf8]{inputenc}
\usepackage[T1]{fontenc}

\usepackage{xcolor}
\usepackage{nameref}
\usepackage{glossaries}
\usepackage{siunitx}
\usepackage{csvsimple}
\usepackage{import}
\usepackage[capitalize]{cleveref}
\usepackage{pgf}
\usepackage{ragged2e}
\usepackage{tabularx}
\usepackage{booktabs}
\usepackage{multirow}
\usepackage{subfig}
\usepackage{tikz}
\usepackage{atbegshi}

\glsdisablehyper

\newcommand{\rnum}[1]{\num[round-mode=figures,round-precision=4]{#1}}
\newenvironment{DIFnomarkup}{}{}

\title{Estimating Motor Symptom Presence and Severity in Parkinson’s Disease from Wrist Accelerometer Time Series using ROCKET and InceptionTime}

\author[1,*]{Cedric Donie}
\author[1]{Neha Das}
\author[1,3]{Satoshi Endo}
\author[1,2,3]{Sandra Hirche}
\affil[1]{Technical University of Munich,
Munich, Germany; TUM School of Computation, Information and Technology, Department of Computer Engineering, Chair of Information-oriented Control}
\affil[2]{Munich Data Science Institute (MDSI)}
\affil[3]{Munich Institute of Robotics and Machine Intelligence (MIRMI)}

\affil[*]{cedric.donie@tum.de}

\newacronym{pd}{PD}{Parkinson's disease}
\newacronym{ld}{L-DOPA}{levodopa}
\newacronym{pddb}{PDDB}{Parkinson's Disease Digital Biomarker}
\newacronym{dream}{DREAM}{Dialogue on Reverse Engineering Assessment and Methods}
\newacronym{aupr}{AUPR}{area under precision-recall curve}
\newacronym{auroc}{AUROC}{area under receiver operating curve}
\newacronym{ap}{AP}{average precision}
\newacronym{map}{mAP}{mean average precision}
\newacronym{api}{API}{application programming interface}
\newacronym{rocket}{ROCKET}{RandOm Convolutional KErnel Transform}
\newacronym{hivecote}{HIVE-COTE}{Hierarchical Vote Collective of Transformation-Based Ensembles}
\newacronym{mjff}{MJFF}{Michael J. Fox Foundation}
\newacronym{tsc}{TSC}{time series classification}
\newacronym{cnn}{CNN}{convolutional neural network}
\newacronym{dnn}{DNN}{dynamic neural network}
\newacronym{svm}{SVM}{support-vector machine}
\newacronym{gmm}{GMM}{Gaussian mixture model}
\newacronym{fog}{FoG}{freezing of gait}
\newacronym{lstm}{LSTM}{long short-term memory network}
\newacronym{sgd}{SGD}{stochastic gradient descent}
\newacronym{mlp}{MLP}{multi-layer perceptron}
\newacronym{aso}{ASO}{almost stochastic order}
\newacronym{tp}{TP}{true positive}
\newacronym{fp}{FP}{false positive}
\newacronym{fn}{FN}{false negative}
\newacronym{tn}{TN}{true negative}
\newacronym{p}{P}{positive}
\newacronym{n}{N}{negative}
\newacronym{ucr}{UCR}{University of California, Riverside}
\newacronym{cote}{COTE}{Collective of Transformation-Based Ensembles}
\newacronym{ba}{BA}{balanced accuracy}
\newacronym{adl}{ADL}{activities of daily living}
\newacronym{cdf}{CDF}{cumulative distribution function}
\newacronym{mdsupdrs}{MDS-UPDRS}{Movement Disorders Society Unified Parkinson’s Disease Rating Scale}
\newacronym{mamae}{MAMAE}{macro-averaged mean absolute error}
\newacronym{psd}{PSD}{power spectral density}


\begin{abstract}
    \gls{pd} is a neurodegenerative condition characterized by frequently changing motor symptoms, necessitating continuous symptom monitoring for more targeted treatment.
    Classical time series classification and deep learning techniques have demonstrated limited efficacy in monitoring \gls{pd} symptoms using wearable accelerometer data due to complex \gls{pd} movement patterns and the small size of available datasets.
    We investigate InceptionTime and \gls{rocket} as they are promising for \gls{pd} symptom monitoring.
    InceptionTime's high learning capacity is well-suited to modeling complex movement patterns, while \gls{rocket} is suited to small datasets.
    With random search methodology, we identify the highest-scoring InceptionTime architecture and compare its performance to \gls{rocket} with a ridge classifier and a  \gls{mlp} on wrist motion data from \gls{pd} patients.
	Our findings indicate that all approaches can learn to estimate tremor severity and bradykinesia presence with moderate performance but encounter challenges in detecting dyskinesia.
	Among the presented approaches, \gls{rocket} demonstrates higher scores in identifying dyskinesia, whereas InceptionTime exhibits slightly better performance in tremor and bradykinesia estimation.
    Notably, both methods outperform the multi-layer perceptron.
    In conclusion, InceptionTime can classify complex wrist motion time series and holds potential for continuous symptom monitoring in \gls{pd} with further development.
    \glsresetall{}
\end{abstract}
\begin{document}
\flushbottom
\maketitle
\thispagestyle{empty}
\begin{tikzpicture} [remember picture, overlay]
\node [yshift=12mm, text width=\textwidth, rectangle, fill=yellow] at (current page.south) {%
\sffamily\small This version of the article has been accepted for publication, after peer review (when applicable) but is not the Version of Record and does not reflect post-acceptance improvements, or any corrections. The Version of Record is available online at: 
\url{https://doi.org/10.1038/s41598-025-04263-2}.%
};
\end{tikzpicture}
\begin{tikzpicture} [remember picture, overlay]
\node [yshift=-12mm,text width=\textwidth] at (current page.north) {\sffamily Scientific Reports volume 15, Article number: 19140, Published: 31 May 2025};
\end{tikzpicture}


\section*{Introduction}
\Gls{pd} is a neurodegenerative condition that severely diminishes patient quality of life.
The predominant physical manifestations of \gls{pd} include \emph{tremor}, \emph{bradykinesia},  and \emph{dyskinesia}, among others~\cite{2018pdbookintro}.
Tremor is described as a trembling of the affected limb that can occur at rest and during activity.
Bradykinesia entails slowed movement.
Dyskinesia is an involuntary twitching and writhing movement.
Because the prevalence of \gls{pd} increases with age, impacting up to \SI{1}{\percent} of individuals over 60~\cite{delau2016pdepi}, the disease poses a growing economic and social problem~\cite{chaudhuri2024}.
Although \gls{ld} and similar dopaminergic medications can alleviate \gls{pd} symptoms, they may also cause the side effect of dyskinesia~\cite{2018pdbookintro}.
Currently, \gls{pd} symptoms are only monitored every three to 12 months~\cite{ng71} during physician visits.
However, symptoms may change throughout the day, e.g., as the medication wears off or its uptake is altered by diet and disease progression.
This infrequent monitoring of rapidly changing symptom severity complicates the selection of interventions that balance symptom relief against dyskinesia.
Thus, methods for continuously and automatically monitoring \gls{pd} would improve intervention and patient quality of life.

A frequently studied method for continuous \gls{pd} monitoring involves detecting symptom severity using unobtrusive, low-cost wearable sensors, such as smartwatch accelerometers~\cite{sigchaDeepLearningWearable2023}.
These inertial sensors provide acceleration components as a multivariate time series, sometimes including additional gyroscope and magnetometer measurements.
Among the various studies that developed models to estimate \gls{pd} movement symptom severity using wearables, one research strand has focused on explicitly modeling (i.e., handcrafting) a fixed set of features from the inertial sensor data prior to conducting data-driven analysis.
For example, Gaussian processes successfully estimated dyskinesia and bradykinesia using wavelet-based features~\cite{endo2018gp}.
Key findings are that modeling \gls{pd} as a dynamical system, with current symptoms depending on past symptoms, is advantageous and that maximum spectral power is useful for the detection of \gls{pd}.
Other studies extracted features such as acceleration variance and channel-delay correlation matrix eigenvalues, applying a \gls{gmm} classifier\cite{williamson2021pdwristworn}; this supports the effectiveness of combining temporal and spectral features.
Conversely, simpler statistical features, such as root-mean-square, were successful with a \gls{svm} for bradykinesia detection~\cite{pastorino2011bradymultiparam}.
Additional proposals include signal processing methods, using dominant pole frequency and amplitude for tremor detection or low-pass filters with explicitly modeled features for bradykinesia~\cite{salarian2007pdambulatory}.
Such features have also been used in conjunction with dynamic neural networks, \glspl{svm}, and hidden Markov models to measure tremor and dyskinesia~\cite{cole2015wearablepd}.
Consequently, explicit features tend to use frequency analysis, but additional modeling of temporal dependencies seems beneficial.

Another research strand employed deep learning for motor symptom severity estimation, leveraging implicit~\cite{bengio2009deeparchai} feature extraction and eliminating the need for fixed features.
\Glspl{cnn} have been shown to detect bradykinesia with greater accuracy than fully connected neural networks, \glspl{svm}, a rule-based classifier (i.e., PART), and AdaBoost in a study of 10 patients; this method operated directly on the raw data and added noise as data augmentation~\cite{eskofierRecentMachineLearning2016}.
Deep neural network architectures designed to better model temporal correlations, such as \glspl{lstm}, have been used for \gls{pd} detection from speech signals~\cite{rizvi2020lstm} and gait data~\cite{balaji2021automatic}. The latter hypothesize and show that \gls{lstm} can learn long-term temporal correlations directly from raw data.
\Gls{cnn}, while typically used for image processing, has also been shown to outperform~\cite{shiranthikaHumanActivityRecognition2020} \glspl{lstm} or perform comparably~\cite{khatun_deep_2022} in human activity recognition from wearable sensors; this is pertinent since activity recognition is akin to \gls{pd} detection.
These empirical results showing that \gls{cnn} are good at \gls{pd} motion detection seem plausible given the close mathematical relationship between successful explicit features (wavelet, Fourier\cite{ronald_bracewell_fourier_2000}, channel-delay correlation) and convolutions.

Nevertheless, all of these approaches suffer from the problems inherent to \gls{pd} detection from inertial sensor data.
The movement patterns characterizing \gls{pd} are complex because the accelerations caused by the symptoms are superimposed on accelerations from \gls{adl}.
These non-symptomatic accelerations vary widely between patients, activities, and the exact sensor positioning.
Furthermore, it is challenging to collect clinical data at scale, and datasets are rather small.
Thus, continuous symptom monitoring necessitates an approach capable of detecting complex patterns in noisy signals with limited training data.
This approach can be reframed as a time series classification task.
Two recently proposed methods for time series classification offer new possibilities and address the problem of complex patterns with limited training data from different perspectives: \emph{InceptionTime}~\cite{fawaz2020inceptiontime} and \gls{rocket}~\cite{dempster2020rocket} have been demonstrated to outperform previous state-of-the-art methods, including \glspl{cnn} and \gls{cnn}-\gls{lstm} combinations (e.g., TapNet~\cite{zhang_tapnet_2020}) on a benchmark consisting of 26 multivariate time series from various domains~\cite{ruiz2020bakeoff}.

InceptionTime and ROCKET have been shown to work for \gls{pd} estimation in detecting overall \gls{pd} presence from eye tracking position and velocity~\cite{uribarriDeepLearningTime2023}, but this differs from our problem of detecting multiple \gls{pd} motor symptom severities from acceleration data.
More closely related work detected \gls{fog} from full-body accelerometer measurements with InceptionTime and the \gls{rocket}-derived MiniROCKET~\cite{klaverComparisonStateoftheartDeep2023,dempsterMiniRocketVeryFast2021}.
InceptionTime has also been used for detecting \gls{pd} from gait through a force-sensitive insole~\cite{zhouDeepLearningBasedClassification2023}, but other approaches, such as ResNet, performed better. Another interesting related work used InceptionTime for detecting \gls{pd} from voice waveforms with a novel generative-adversarial technique and found that reducing InceptionTime's model complexity and number of parameters led to higher performance, with InceptionTime ranking second-best behind the ResNet \gls{cnn} variant\cite{rey-paredesTimeSeriesClassification2025}.
However, existing work using InceptionTime and ROCKET do not provide multi-level symptom severity classification \cite{uribarriDeepLearningTime2023, klaverComparisonStateoftheartDeep2023} and restrict the prediction to the presence or absence of \gls{pd}. Furthermore, they often either use research-grade \cite{uribarriDeepLearningTime2023, zhouDeepLearningBasedClassification2023, rey-paredesTimeSeriesClassification2025} or full-body\cite{klaverComparisonStateoftheartDeep2023} sensors.

\begin{figure}
    \setlength{\parindent}{0pt}
    \begin{minipage}{0.48\textwidth}
        \centering\noindent
        \includegraphics[scale=0.635]{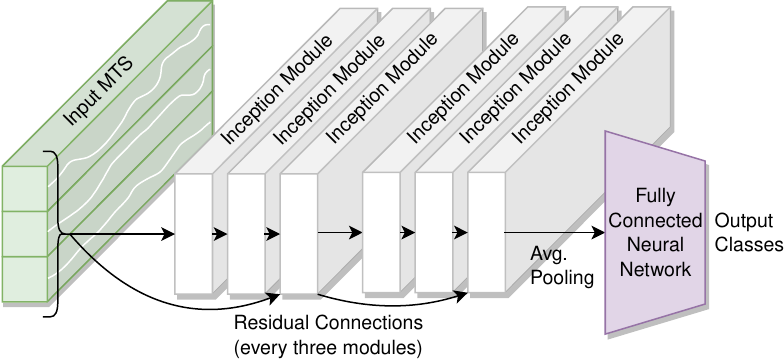}
        \sffamily
        \vspace*{8mm}
        \footnotesize
        (a) Inception Network
    \end{minipage}
    \hfill
    \begin{minipage}{0.48\textwidth}
        \centering\noindent
        \includegraphics[scale=0.635]{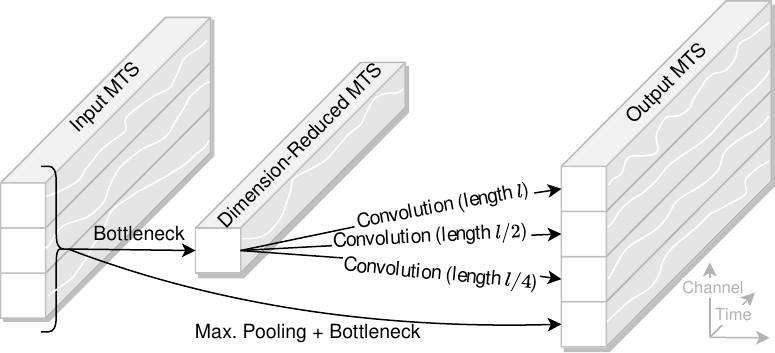}
        \sffamily
        \vspace*{8mm}
        \footnotesize
        (b) Inception Module
    \end{minipage}
    \caption{(a) Simplified depiction (inspired by the seminal work~\cite{fawaz2020inceptiontime}) of an Inception network consisting of Inception modules, average pooling, and a fully connected neural network to generate class prediction from an input multivariate times series (MTS).
    The network's depth is six, equal to the number of Inception modules. \\
    (b) The Inception module's bottleneck first reduces the input MTS to a univariate time series, and then convolutional filters are applied along the time axis. Additionally, the result of maximum pooling and a bottleneck is concatenated directly to the output. In this example, the Inception module takes a three-dimensional MTS as input and outputs a four-dimensional MTS. The module has three filters and a filter length of $l$.
    }
    \label{fig:inceptiontimearch}
\end{figure}

\begin{figure}
    \centering
    \includegraphics[scale=0.635]{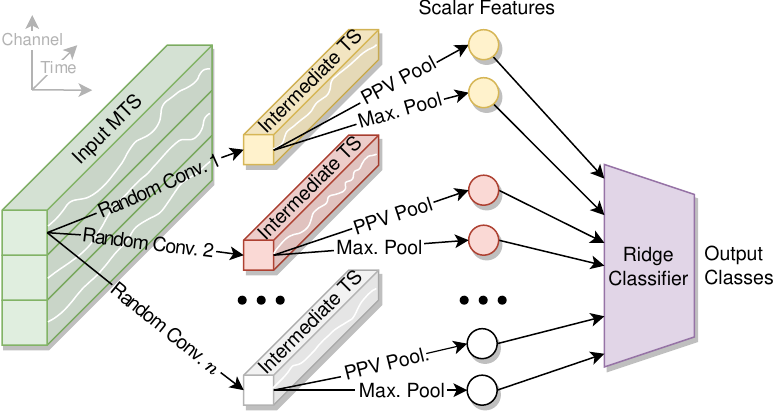}
    \caption{Simplified depiction of ROCKET~\cite{dempster2020rocket}. Random convolutional kernels are applied to every time series of the input MTS (shown for the first time series only), yielding an intermediate time series. Proportion of positive values (PPV) and maximum pooling extract two scalar features per intermediate time series, which are inputs to a ridge classifier. The depicted example has $n$ random kernels and produces $2n$ random scalar features per input dimension.}
    \label{fig:rocketarch}
\end{figure}

InceptionTime is an ensemble of five \emph{Inception networks} with different random initializations.
Each Inception network comprises at least one \emph{Inception module}, followed by global average pooling, and then fully connected layers to generate class predictions, as shown in \cref{fig:inceptiontimearch}\cite{fawaz2020inceptiontime}.
Inception modules transform multivariate time series using convolution filters, maximum pooling, and concatenation.
The convolutional filters have varying lengths, allowing the Inception modules to learn long-duration features from time series.
By stacking Inception modules, which may include long filters, InceptionTime can have a very large receptive field.
Ensembling multiple Inception networks reduces variability for more consistent performance.
We hypothesize that this expansive receptive field will enable InceptionTime to effectively learn the complex patterns present in \gls{pd} time series data; potentially, it may even compete with \gls{lstm} on learning long-term features.
The parameter sharing inherent to convolutions will be well-suited to scenarios with limited data.
\Gls{rocket} uses convolutional kernels similar to those in \glspl{cnn}~\cite{dempster2020rocket}.
Instead of learning the kernel parameters from the training data, \gls{rocket} generates thousands of random kernels to create features.
It then learns only a small set of linear weights from the training data using ridge regression, as shown in \cref{fig:rocketarch}.
Using random rather than trained kernels makes \gls{rocket} well-suited for \gls{pd} symptom severity estimation with limited training data.
It is well known that as variance decreases with more model parameters, bias increases; random kernels reduce the number of parameters drastically and may result in a more favorable ratio of parameters to data points.
As \gls{pd} affects frequency-related features, we expect that the convolutions intrinsic to both methods will facilitate \gls{pd} severity estimation.

Despite InceptionTime's capacity to capture complex patterns and \gls{rocket}'s effectiveness with limited data, to the best of our knowledge, neither approach has been studied for detecting tremor, bradykinesia, and dyskinesia from wearable wrist accelerometers.
The present work addresses this gap by systematically evaluating the performance, calibration, and hyperparameters of InceptionTime and \gls{rocket} for \gls{pd} motor symptom severity estimation.
Our rigorous analysis highlights the advantages and disadvantages depending on the symptom of interest, desired robustness, and target outcomes.
We demonstrate that both InceptionTime and \gls{rocket} predict symptoms from accelerometer time series during \gls{adl} more effectively than explicitly modeled wavelet-derived features.
Although these methods do not reach clinically relevant performance, this study reveals the strengths and limitations of two recently introduced time series classification approaches, \gls{rocket} and InceptionTime, for \gls{pd} symptom severity estimation.
Our results show that the presented methods, with the default hyperparameters, can be used as a baseline for future research investigating \gls{pd} symptom severity estimation.

\section*{Methods}
We compare the InceptionTime ensemble and \gls{rocket} against a baseline feature-based classifier for estimating the severity or presence/absence of tremors, bradykinesia, and dyskinesia from the patient's smartwatch.
\subsection*{Data}
\label{sec:data}
The dataset is a subset of the publicly available 2021 \gls{mjff} Levodopa Response Study~\cite{daneault2021mjffldopa1,vergaradiaz2021mjffldopa2} repository.

\subsubsection*{Particpants and Demographics}
\newcommand{\hyscoremean}{2.269230769230769}
\newcommand{\agemean}{62.92307692307692}
\newcommand{\agestd}{8.629823066195863}
\newcommand{\agemin}{46}
\newcommand{\agemax}{80}
\newcommand{\malefrac}{70.37037037037037}
\newcommand{\femalefrac}{29.629629629629626}
\newcommand{\rightfrac}{96.29629629629629}
\newcommand{\leftfrac}{3.7037037037037033}
\newcommand{\rightmostaffectedfrac}{66.66666666666666}
\newcommand{\leftmostaffectedfrac}{25.925925925925924}
\newcommand{\bilateralmostaffectedfrac}{7.4074074074074066}
\newcommand{\hyIIcount}{21}
\newcommand{\hyIIIcount}{3}
\newcommand{\hyIVcount}{2}
\newcommand{\timesinceldopamedian}{210.0}
\newcommand{\timesinceldopamean}{168.26086955}
\newcommand{\timesinceldopaQone}{32.5}
\newcommand{\timesinceldopaQthree}{257.0}
\newcommand{\durationsum}{162.0450166598956}
{\sisetup{round-mode=figures,round-precision=3,range-phrase=--}%
\DeclareSIUnit\year{yr}
The \gls{mjff}'s inclusion criteria were community-dwelling, \gls{pd} diagnosis, 30 to 80 years of age, taking \gls{ld} at the time of data collection, self-reported motor fluctuations, self-reported dyskinesia (at least mild), and ability to operate a smartphone~\cite{daneault2021mjffldopa1}.
Patients with other severe neurological issues or deep brain stimulation were excluded~\cite{daneault2021mjffldopa1}.
The selected subset includes data from 27 patients.
Patients were \agemin{} to \agemax{} years of age (mean \SI{\agemean}{\year}, S.D. \SI{\agestd}{\year}) with
Hohn and Yahr scores of II (\hyIIcount{} patients), III (\hyIIIcount{} patients), IV (\hyIVcount{} patients), and unknown (1 patient).
\SI{\malefrac}{\percent} were male and \SI{\femalefrac}{\percent} were female gender.
\Gls{pd} motor symptoms affected \SI{\bilateralmostaffectedfrac}{\percent} equally on both sides of the body, \SI{\rightmostaffectedfrac}{\percent} more on the right, and the rest on the left.
\SI{\rightfrac}{\percent} were right-handed and \SI{\leftfrac}{\percent} were left-handed.%

\subsubsection*{Collected Measurements}
Patients were monitored in a clinic on days one and four of the study.
On day one, patients arrived in an on-medication state, having taken \gls{ld} on their regular schedule, but on day four, patients arrived at least 12 hours after the last \gls{ld} dose.
The 27 patients wore a GENEActiv smartwatch on the most affected limb as they performed a battery of pre-defined motor tasks (such as standing, walking, and typing) in a laboratory setting~\cite{daneault2021mjffldopa1}.
The test battery was repeated six to eight times.
On day one, the median time between the last \gls{ld} dose and the motor tasks was \SI{\timesinceldopamedian}{\minute} (IQR \SIrange{\timesinceldopaQone}{\timesinceldopaQthree}{\minute}).
The GENEActiv measures acceleration in three dimensions at \SI{50}{\hertz}, producing multiple multivariate time series per patient and task.
These time series vary in duration with the time taken for a task and have a mean duration of \SI{29.2}{\second} (S.D. \SI{11.6}{\second}).
}

\subsubsection*{Annotation}
The Levodopa Response Study used several \gls{mdsupdrs} certified clinical researchers during the in-clinic monitoring.
A given task for a given patient was rated according to \gls{mdsupdrs} by a single specialist~\cite{daneault2021mjffldopa1}.
The dataset thus has symptom severity annotations for each of the acceleration time series.
Presence/absence of bradykinesia and dyskinesia are boolean, and tremor is on an ordinal scale from zero (no symptoms) to four (severe symptoms)~\cite{daneault2021mjffldopa1}.
We remove data points with missing or implausible annotations from the dataset.
In total, the dataset contains \SI[round-mode=figures,round-precision=4]{\durationsum}{\hour} annotated GENEActiv acceleration measurements.

\subsection*{Evaluation Design}
Comparing the machine learning approaches fairly requires appropriate data handling and selection of suitable evaluation metrics.
\subsubsection*{Data Split}
\label{sec:datasplit}

We split the pre-processed data into disjoint training, test, and validation sets, ensuring that no patient appears in multiple sets.
Each split resembles the class distribution of the overall dataset through data stratification.
\Cref{tab:datasplit} shows the split between training, validation, and test data.
The total duration refers to the sum of the durations of all tasks with a given symptom annotation.
There is a substantial imbalance in the class labels, with more instances of no (or mild) symptoms than strong symptoms.
\begin{table}
    \caption{Split between training, validation, and test data. Total duration is the sum of the duration of all time series for a given symptom severity. The number of patients refers to the count of patients with at least one task rated at the given symptom severity.}
    \label{tab:datasplit}
    \centering
\begin{DIFnomarkup}
\begin{tabular}{llS[round-mode=figures,round-precision=3]S[round-mode=figures,round-precision=3]S[round-mode=figures,round-precision=3]SSS}
\toprule
           {} &      {} & \multicolumn{3}{c}{{Total Duration /h}} & \multicolumn{3}{c}{{Number of Patients}} \\
  \cmidrule(lr){3-5} \cmidrule(lr){6-8}
  {Symptom} & {Score} &    {Train} &     {Validate} &     {Test} &     {Train} & {Validate} & {Test} \\
  \midrule
  \multirow{3}{*}{bradykinesia} &     n/a &  11.880878 &  0.934911 &   3.527739 &          19 &     2 &      6 \\
                                &      no &  18.341456 &  1.587172 &   5.749772 &          19 &     2 &      6 \\
                                &     yes &   7.743661 &  1.905794 &   2.343622 &          19 &     2 &      6 \\
  \midrule
  \multirow{2}{*}{dyskinesia}   &      no &  33.691450 &  4.034733 &  10.825050 &          19 &     2 &      6 \\
                                &     yes &   4.274544 &  0.393144 &   0.796083 &          13 &     2 &      4 \\
  \midrule
  \multirow{5}{*}{tremor}       &       0 &  26.773978 &  1.415744 &   8.212411 &          19 &     2 &      6 \\
                                &       1 &   8.162694 &  2.070278 &   2.054656 &          14 &     2 &      6 \\
                                &       2 &   2.310700 &  0.682539 &   1.097578 &          10 &     2 &      3 \\
                                &       3 &   0.683383 &  0.259317 &   0.223456 &           4 &     1 &      2 \\
                                &       4 &   0.035239 &  0        &   0.033033 &           2 &     0 &      1 \\
   \bottomrule
\end{tabular}
\end{DIFnomarkup}
\end{table}

For hyperparameter tuning, we hold out the test set for final evaluation and apply grouped stratified cross-validation using the remaining data.~\cite{vablas2019mlvallimiteddata,cawley2010overfitting}.
Specifically, we create five folds such that class proportions remain consistent across folds, and no patient appears in multiple folds.
Five cross-validation folds lead to approximately \SI{80}{\percent} training data and \SI{20}{\percent} validation data for each model.
Scores are averaged across these folds.

\subsubsection*{Statistical Analysis}
\label{sec:stats}
Throughout the present work, we use the metrics of \gls{ba} and \gls{ap}.
Balanced accuracy is the mean of the recall of each class~\cite{mosley2013balanced}.
\Gls{ap}~\eqref{eq:ap} is the mean of the precision $P$ at each threshold $i$, weighted by the change in recall $R$~\cite{wanhua2015prroc}.
Averaging the one-vs.-rest per-class \gls{ap} over all classes yields the  \gls{map}.
\begin{equation}
	\text{AP} = \sum_{i} (R_i - R_{i-1}) P_i \label{eq:ap}
\end{equation}
In addition, we use accuracy, \gls{auroc}, and the F1-score \eqref{eq:f1} for comparison with related studies.
\begin{equation}
    \text{F1} = 2 \cdot \frac{P \cdot R}{P + R} \label{eq:f1}
\end{equation}
For the multiclass task of tremor prediction, F1 is macro-averaged across all classes in a one-vs.-rest fashion.
Multiclass \gls{auroc} is calculated by averaging the scores of all pairwise class combinations.
This one-vs.-one \gls{auroc} is less sensitive to class imbalance than a one-vs.-rest calculation\cite{handSimpleGeneralisationArea2001}.
\Gls{rocket} provides class likelihoods rather than the probabilities required for \gls{auroc} calculation and calibration curve analysis.
Softmax \eqref{eq:softmax} constrains the \gls{rocket} class likelihood vector $\vec{x}=\{x_1, x_2, x_3, \ldots, x_n\}$ to the interval $[0;1]$, thereby approximating probabilities.
\begin{equation}
    \text{softmax}(x_i)=\frac{e^{x_i}}{\sum_{k=1}^n e^{x_k}}\label{eq:softmax}
\end{equation}

Tremors are assessed by clinicians on an ordinal scale, and strongly misclassifying tremor severity is worse than a prediction that is off by one.
To account for class imbalance and the degree of misclassification, we use the \gls{mamae}, which is the macro-average of the per-class mean absolute error\cite{baccianellaEvaluationMeasuresOrdinal2009}. \Gls{mamae} is calculated according to \cref{eq:mamae}, where $N$ is the number of test samples, $C$ is the set of groundtruth classes, $y_i$ is the groundtruth class of the $i$-th test sample, and $\hat{y_i}$ is the predicted class.
\begin{align}
    \text{MAMAE} &= \frac{1}{N} \sum_{c \in C} \frac{\sum_{i=1}^N \delta_{c,y_i} \left| \hat{y_i} - y_i \right|}{\sum_{i=1}^N\delta_{c,y_i}} \label{eq:mamae}\\
    \delta_{i, j}&= \begin{cases}0 & \text { if } i \neq j \\ 1 & \text { if } i=j\end{cases} \nonumber
\end{align}

To judge how well the model's predicted probability matches the actual occurrence probability, reliability plots (also known as calibration curves) can be used.
To compare the classifiers' predicted probability to the actual probability, the latter value must be determined; this is non-trivial because the presence/absence of a symptom is binary. Thus, patients with similar predicted symptom probabilities must be grouped together.
We use the \textit{SmoothECE} approach, which avoids the pitfalls of arbitrarily binning by predicted probabilities or manually choosing smoothing parameters~\cite{blasiokSmoothECEPrincipled2023}.
SmoothECE regresses the expected value of the conditional relative occurrence frequency $E[y|f]$ against the predicted probability $f$ via Gaussian kernel smoothing~\cite{blasiokSmoothECEPrincipled2023}.
The SmoothECE metric reflects the deviation from an ideal calibration curve, with lower values indicating better calibration.
Crucially, the kernel bandwidth is determined automatically, eliminating all parameter tuning and enhancing comparability with future studies.

Comparing deep learning models with statistical rigor is challenging because conventional statistical tests often assume normally distributed results, which may not hold~\cite{dror2019deepdominance}.
Instead of a conventional test, we apply \emph{\gls{aso}}~\cite{dror2019deepdominance,ulmer2022deepsignificance} to the scores of multiple training runs for each model and run \num{1000} bootstrap iterations.
\Gls{aso} extends the concept of \emph{stochastic dominance}, whereby an algorithm A is stochastically dominant over algorithm B if and only if the empirical \gls{cdf} of A's scores is always greater than the \gls{cdf} of B's scores~\cite{dror2019deepdominance}.
\Gls{aso} allows stochastic dominance---which is too restrictive for practical purposes---to be violated to a degree of $\epsilon_\text{min}$, estimating the upper bound of $\epsilon_\text{min}$ via bootstrapping~\cite{ulmer2022deepsignificance}.
We train each model ten times and set significance level $\alpha = 0.05$, suggesting that ten samples are sufficient for statistical comparison in nearly all cases.
The present work henceforth considers A stochastically dominant over B for $\epsilon_\text{min} < 0.2$ with 1000 bootstrap iterations, as suggested by~\cite{ulmer2022deepsignificance}.
Bootstrap power analysis of model scores~\cite{yuan2003bootstrappoweranalysis} with \num{5000} iterations ensures a power of at least 0.8~\cite{ulmer2022deepsignificance}.
We perform Bonferroni correction for a total of 31 comparisons (based on four models, three symptoms, and two metrics; considering symmetry and insufficient power).

To explain misclassifications, we examine the distribution of performed motor tasks among false positives and false negatives.
The expected count $E_{t,\hat{y}, y}$ for a motor task $t \in T$, predicted class $\hat{y}$, and actual symptom severity class $y$ is computed from the observed count $O$ according to \eqref{eq:expecteddist}, where $C$ denotes the set of severity levels for a given symptom:
\begin{equation}\label{eq:expecteddist}
E\_{t,\hat{y},y} = \frac{\sum_{i \in C} O_{t,i,y}}{\sum_{k \in T} \sum_{i \in C} O_{k,i,y}} \sum_k O_{k,\hat{y},y}
\end{equation}
If the misclassifications by the model were uniformly distributed across motor tasks, the observed and expected counts would closely match.
Deviations from this distribution indicate task-specific error patterns.
Expected and observed frequencies are calculated across all ten training repetitions. 
To address class imbalance arising from the limited number of strong tremor instances, tremor severity was binarized into present (severity $\geq 1$) and absent (severity $=0$).
To enhance statistical robustness, differences between observed and expected counts for a given ground-truth severity and motor task are reported only if the difference is at least \SI{40}{\percent}, with both observed and expected counts consisting of at least ten instances.
Welch's method \cite{welchUseFastFourier1967} is employed for power spectral density (PSD) estimation, enabling the identification of dominant motion frequencies present within the acceleration time series for each motor task.
The complexity of each time series is quantified using \textit{sample entropy}, which is particularly suited to short and noisy physiological data\cite{richmanPhysiologicalTimeseriesAnalysis2000}.
Higher sample entropy values indicate greater complexity or randomness, whereas lower values suggest simpler, more regular patterns.
Finally, the energy of a time series is directly proportional to the square of its amplitudes, and its power is directly proportional to its standard deviation.
\Gls{psd}, standard deviation, and sample entropy are calculated from the magnitude of the acceleration time series.

\subsection*{Model Development}
We aim to optimize InceptionTime and \gls{rocket} \gls{pd} prediction through hyperparameter optimization.
Two InceptionTime models are developed: one with the default hyperparameters and one with extensive tuning.
To evaluate the hypothesis that they \gls{rocket} and InceptionTime will implicitly learn good features, we also develop a baseline classifier using explicit features.
Finally, the developed models are trained on the combined training and validation dataset.
The selected time series classification approaches require equal-length time series, but raw GENEActiv measurements are of unequal length.
Thus, we normalized the data length using a forward-sliding window approach with two hyperparameters: window length and overlap proportion between windows.
We use windows of \SI{30}{\second} length with \SI{50}{\percent} overlap unless stated otherwise.
This fixed window length and overlap mean that more training samples will be generated from longer series.
Preliminary tests indicated improved performance with longer windows and greater overlap until performance plateaus at \SI{50}{\percent} overlap.
Current research lacks consensus on the optimal window length.~\cite{cole2015wearablepd,eskofierRecentMachineLearning2016,pastorino2011bradymultiparam,salarian2007pdambulatory,bikias2021fogdeeplearning}.
Since all examples of tremor severity four in the training data have durations of less than \SI{30}{s}, they are not used for training any of the models.
Consequently, the learned models cannot predict tremor severity four, and we therefore combine severity classes three and four into a single class, referred to as ``3--4''.

\subsubsection*{InceptionTime Hyperparameter Tuning}\label{sec:hyper}
InceptionTime has several hyperparameters\cite{fawaz2020inceptiontime}:
\emph{Filter size} is the length of the longest 1d-convolution filter in the Inception modules (e.g., filter size $l$ in \cref{fig:inceptiontimearch}).
\emph{Number of filters} refers to the number of filters each Inception module contains (e.g., three filters in \cref{fig:inceptiontimearch}).
For example, a filter size of 64 and four filters will result in filters of length 64, 32, 16, and 8 for each Inception module.
\emph{Depth} is the number of stacked Inception modules (e.g., depth 6 in \cref{fig:inceptiontimearch}).
In addition to the filter size, the number of filters, and the depth, we include the window length in the hyperparameter search.
We activate residual connections because they improve accuracy and retain the default batch size of 64 as batch size does not affect accuracy~\cite{fawaz2020inceptiontime}.

Because InceptionTime was originally developed using the \gls{ucr} archive~\cite{fawaz2020inceptiontime}, which contains many different time series (including human activity recognition~\cite{ucrarchive2018}), the default architecture might already be transferable to use cases, such as \gls{pd} symptom severity estimation.
We first train InceptionTime with the default hyperparameters~\cite{fawaz2020inceptiontime}.
Next, selected hyperparameters must be optimized to maximize scores based on at least one metric.
We employ random search due to its efficiency compared to grid search, particularly when certain hyperparameters are more important than others~\cite{bergstra12randomsearch}.
Window length is a real number from \SIrange{3}{30}{\s} with fixed \SI{50}{\percent} overlap.
Filter length is an integer from 8 to 255.
Depth is an integer from 1 to 11.
The number of filters is a power of two from $2^1=2$ to $2^6=64$.
The window length is sampled uniformly with replacement; all other hyperparameters are sampled uniformly without replacement.
We perform 60 random search trials using cross-validation as described above to split the data into training and validation data.
For tremor, dyskinesia, and bradykinesia, we train 300 models each, resulting in as many as 900 models being trained.
Training is terminated after 600 epochs, based on preliminary tests indicating performance plateaus or declines after 600 epochs, and we hope that this incentivizes architectures that are invariant to epoch count.

\subsubsection*{ROCKET Hyperparameter Selection}
\label{sec:rockethyper}
\Gls{rocket} hyperparameters include the number of convolution kernels, classifier choice (ridge or logistic regression), and window length.
This work retains the default random kernel parameters (\num{10000} kernels, see \cref{fig:rocketarch}), which ``form an intrinsic part of ROCKET'' and ``do not need to be `tuned' for new datasets'' because they are optimized for a variety of datasets~\cite{dempster2020rocket}.
We opt for the ridge classifier and tune its regularization strength via cross-validation.

\subsubsection*{Wavelet-Based Feature Engineering and Classifier Development}
The baseline classifier is a \gls{mlp} applied to 70 wavelet-based features.
Our baseline deliberately uses a simple, generic classifier combined with wavelet features that capture domain-specific knowledge as a counterpoint to the other black-box methods with learned features.
As InceptionTime uses an \gls{mlp} and \gls{rocket} uses a ridge classifier in the final stage, using wavelets with an \gls{mlp} allows us to compare the efficacy of wavelets-derived features, identified as state-of-the-art for \gls{pd} classification in prior studies~\cite{langMultiLayerGaussianProcess2019,endo2018gp,wagnerWaveletbasedApproachMonitoring2017,vimalajeewaParkinsonsDiseaseDiagnosis2023}, to InceptionTime and \gls{rocket}'s implicit feature extraction.
More concretely, wavelet features encode the domain knowledge that different symptom severities will affect the wrist motion frequency and have been used in prior studies together with Gaussian processes and \gls{svm}.
We derive these 70 features by determining the root-mean-square, standard deviation, maximum, kurtosis, skew, power spectral distribution maximum, and power spectral distribution minimum for nine levels of wavelet decomposition and the original signal~\cite{endo2018gp}.
In contrast to the convolutions used by the first stages InceptionTime and \gls{rocket}, an \gls{mlp} is not affected by the order of features.
The \gls{mlp} is realized with two hidden layers of 128 sigmoid neurons each because a grid search on \gls{pd} data revealed that this represents the ideal topology.
We minimize categorical cross-entropy loss with the parameter optimizer Adam~\cite{kingma2014adam}.%
The categorical cross-entropy loss for an $N$-class prediction problem is calculated in \cref{eq:categoricalcrossentropy} from the probability prediction $\hat{y}$ of the classifier and the one-hot encoded labels $y$.
\begin{equation}\label{eq:categoricalcrossentropy}
    L = - \sum_{i=0}^N y_i \cdot \log\hat{y}_i
\end{equation}

\section*{Results}
\label{sec:overallresults}

This section first describes the results of hyperparameter tuning and models that are selected for the final evaluation.
We then report the results of the final models on the test dataset.

\subsection*{Selected Models and Hyperparameters}\label{sec:hyperresults}
During the hyperparameter tuning process, the test set is not used.

\subsubsection*{InceptionTime}
As \gls{map} and \gls{ap} are positively correlated with balanced accuracy, \gls{ap} is used for model selection henceforth.
For tremor, the \gls{map} has a moderate positive correlation with the window length, as shown in \cref{fig:hyperscatter}.
The bradykinesia and dyskinesia \gls{ap} have a weak positive correlation with the window length.
All other hyperparameters have negligible impact on the \gls{map} or \gls{ap}, except for very slightly decreasing \gls{map} with increasing filter length for tremor.
\begin{figure*}
    \centering
        {
        \renewcommand{\rmfamily}{\sffamily}
        \sisetup{round-mode=uncertainty,round-precision=2}
        \setlength{\tabcolsep}{0.6mm}
        \import{figures/}{hypertuningscatter.pgf}
    }
    \caption{Average precision (AP) and mean AP (mAP) in relation to InceptionTime hyperparameters.
    Network depth, filter length, and the number of filters do not affect the AP. 
    The mAP and AP increase with increasing window length. Spearman's Rank Correlation Coefficient is denoted by $r_s$.
    The arrow denotes the best hyperparameters; the scores and standard deviation are in the rightmost column.
    }
    \label{fig:hyperscatter}
\end{figure*}
Note that the \gls{ap} scores for dyskinesia are low, and the balanced accuracy is often close to the balanced accuracy of 0.5 expected from random classifiers.
The standard deviations of \gls{map} or \gls{ap} and balanced accuracy are considerable, and the interval of $\pm$\,S.D. around the scores could encompass many of the architectures with lower mean scores.

\subsubsection*{ROCKET}
Longer windows should lead to higher performance because \gls{rocket} excels with little training data~\cite{dempster2020rocket} while benefiting from the larger vector during inference, as the following experiments confirm:
For tremor, we find a \gls{map} of 0.404 with \SI{5}{\s} windows, 0.457 with \SI{15}{\s}, and 0.565 with \SI{30}{\s}.
A similar pattern is observed for bradykinesia (\gls{ap} of 0.681 with \SI{5}{\s}, 0.712 with \SI{15}{\s}, and 0.727 with \SI{30}{s}) and for dyskinesia (\gls{ap} of 0.119 with \SI{5}{\s}, 0.100 with \SI{15}{\s}, and 0.140 with \SI{30}{s}).
Thus, we use \SI{30}{\s} windows.

\newcommand{\epstremorAPgpoverrocket}{\rnum{0.9956257008172568}}
\newcommand{\epstremorAPgpoverinceptiontime}{\rnum{1.0}}
\newcommand{\epstremorAPgpoverinceptiontimetuned}{\rnum{0.9999858390040073}}
\newcommand{\epstremorAProcketovergp}{\rnum{0.0}}
\newcommand{\epstremorAProcketoverinceptiontime}{\rnum{0.006780333009270851}}
\newcommand{\epstremorAProcketoverinceptiontimetuned}{\rnum{0.06151277006727887}}
\newcommand{\epstremorAPinceptiontimeovergp}{\rnum{0.0}}
\newcommand{\epstremorAPinceptiontimeoverrocket}{\rnum{0.9943061838369277}}
\newcommand{\epstremorAPinceptiontimeoverinceptiontimetuned}{\rnum{0.9840556632013893}}
\newcommand{\epstremorAPinceptiontimetunedovergp}{\rnum{0.0}}
\newcommand{\epstremorAPinceptiontimetunedoverrocket}{\rnum{1.0}}
\newcommand{\epstremorAPinceptiontimetunedoverinceptiontime}{\rnum{1.0}}
\newcommand{\epstremorBAgpoverrocket}{\rnum{0.9949778372437125}}
\newcommand{\epstremorBAgpoverinceptiontime}{\rnum{0.9968668857583824}}
\newcommand{\epstremorBAgpoverinceptiontimetuned}{\rnum{1.0}}
\newcommand{\epstremorBArocketovergp}{\rnum{0.0}}
\newcommand{\epstremorBArocketoverinceptiontime}{\rnum{0.025477662944362107}}
\newcommand{\epstremorBArocketoverinceptiontimetuned}{\rnum{0.00010449314057421579}}
\newcommand{\epstremorBAinceptiontimeovergp}{\rnum{0.0}}
\newcommand{\epstremorBAinceptiontimeoverrocket}{\rnum{0.997539614332175}}
\newcommand{\epstremorBAinceptiontimeoverinceptiontimetuned}{\rnum{0.18644705211093834}}
\newcommand{\epstremorBAinceptiontimetunedovergp}{\rnum{0.0}}
\newcommand{\epstremorBAinceptiontimetunedoverrocket}{\rnum{0.9918275044331996}}
\newcommand{\epstremorBAinceptiontimetunedoverinceptiontime}{\rnum{1.0}}
\newcommand{\epsdyskinesiaAPgpoverrocket}{\rnum{0.997234301992298}}
\newcommand{\powerdyskinesiaAPinceptiontime}{\rnum{0.2944}}
\newcommand{\powerdyskinesiaAPinceptiontimetuned}{\rnum{0.4092}}
\newcommand{\epsdyskinesiaAProcketovergp}{\rnum{0.0}}
\newcommand{\epsdyskinesiaBAgpoverrocket}{\rnum{1.0}}
\newcommand{\epsdyskinesiaBAgpoverinceptiontime}{\rnum{1.0}}
\newcommand{\epsdyskinesiaBAgpoverinceptiontimetuned}{\rnum{1.0}}
\newcommand{\epsdyskinesiaBArocketovergp}{\rnum{0.252722082691886}}
\newcommand{\epsdyskinesiaBArocketoverinceptiontime}{\rnum{1.0}}
\newcommand{\epsdyskinesiaBArocketoverinceptiontimetuned}{\rnum{1.0}}
\newcommand{\epsdyskinesiaBAinceptiontimeovergp}{\rnum{0.042712060559598385}}
\newcommand{\epsdyskinesiaBAinceptiontimeoverrocket}{\rnum{0.12028023050222077}}
\newcommand{\epsdyskinesiaBAinceptiontimeoverinceptiontimetuned}{\rnum{1.0}}
\newcommand{\epsdyskinesiaBAinceptiontimetunedovergp}{\rnum{0.05776493525767466}}
\newcommand{\epsdyskinesiaBAinceptiontimetunedoverrocket}{\rnum{0.16783023373902262}}
\newcommand{\epsdyskinesiaBAinceptiontimetunedoverinceptiontime}{\rnum{1.0}}
\newcommand{\epsbradykinesiaAPgpoverrocket}{\rnum{1.0}}
\newcommand{\epsbradykinesiaAPgpoverinceptiontime}{\rnum{0.9975915738807771}}
\newcommand{\epsbradykinesiaAPgpoverinceptiontimetuned}{\rnum{0.9979510236653434}}
\newcommand{\epsbradykinesiaAProcketovergp}{\rnum{0.5967803669895038}}
\newcommand{\epsbradykinesiaAProcketoverinceptiontime}{\rnum{0.9984992860961188}}
\newcommand{\epsbradykinesiaAProcketoverinceptiontimetuned}{\rnum{0.9988671511439497}}
\newcommand{\epsbradykinesiaAPinceptiontimeovergp}{\rnum{0.0}}
\newcommand{\epsbradykinesiaAPinceptiontimeoverrocket}{\rnum{0.0}}
\newcommand{\epsbradykinesiaAPinceptiontimeoverinceptiontimetuned}{\rnum{0.27852706184574594}}
\newcommand{\epsbradykinesiaAPinceptiontimetunedovergp}{\rnum{0.0}}
\newcommand{\epsbradykinesiaAPinceptiontimetunedoverrocket}{\rnum{0.0}}
\newcommand{\epsbradykinesiaAPinceptiontimetunedoverinceptiontime}{\rnum{1.0}}
\newcommand{\epsbradykinesiaBAgpoverrocket}{\rnum{0.9985711661581307}}
\newcommand{\epsbradykinesiaBAgpoverinceptiontime}{\rnum{0.9961689111733845}}
\newcommand{\epsbradykinesiaBAgpoverinceptiontimetuned}{\rnum{1.0}}
\newcommand{\epsbradykinesiaBArocketovergp}{\rnum{0.0009138367842256707}}
\newcommand{\epsbradykinesiaBArocketoverinceptiontime}{\rnum{0.9970970252687779}}
\newcommand{\epsbradykinesiaBArocketoverinceptiontimetuned}{\rnum{1.0}}
\newcommand{\epsbradykinesiaBAinceptiontimeovergp}{\rnum{0.0}}
\newcommand{\epsbradykinesiaBAinceptiontimeoverrocket}{\rnum{0.0}}
\newcommand{\epsbradykinesiaBAinceptiontimeoverinceptiontimetuned}{\rnum{0.769635663149087}}
\newcommand{\epsbradykinesiaBAinceptiontimetunedovergp}{\rnum{0.0}}
\newcommand{\epsbradykinesiaBAinceptiontimetunedoverrocket}{\rnum{0.08870933787428194}}
\newcommand{\epsbradykinesiaBAinceptiontimetunedoverinceptiontime}{\rnum{1.0}}
\subsection*{Final Model Performance}\label{sec:resultsga}\renewcommand{\rnum}[1]{\num[round-mode=figures,round-precision=3]{#1}}
InceptionTime and \gls{rocket} are always stochastically dominant over the wavelet \gls{mlp}.
In none of the studied cases does hyperparameter tuning yield superior performance over the default InceptionTime hyperparameters.
The performance of InceptionTime compared to \gls{rocket} varies by symptom and metric.
Statistical power exceeds 0.8 unless stated otherwise.

\subsubsection*{Tremor}
\label{sec:resultstremor}
All classifiers perform better than random classifiers, as shown in \cref{fig:genactivmodelcomp}.
Hyperparameter tuning does not improve InceptionTime scores.
Notably, variability between training runs is very high for InceptionTime, especially after hyperparameter tuning, but low for \gls{rocket}.
\begin{figure}
    \renewcommand{\rmfamily}{\sffamily}
    \centering
    \import{figures/}{modelcomparison.pgf}
    \caption{Comparison of all classifiers for smartwatch wrist sensor data. Each classifier is trained and evaluated ten times. The dashed line represents the scores expected from random classifiers. The whiskers extend to the furthest data point to a maximum of 1.5 interquartile ranges past the first and third quartile. Diamonds represent outliers not within the whiskers.}
    \label{fig:genactivmodelcomp}
\end{figure}
\Gls{rocket} has higher scores than default InceptionTime ($\epsilon_\text{min,mAP}=\epstremorAProcketoverinceptiontime$, $\epsilon_\text{min,BA}=\epstremorBArocketoverinceptiontime$).
\Gls{rocket} scores higher than the hyperparameter-tuned InceptionTime ($\epsilon_\text{min,mAP}=\epstremorAProcketoverinceptiontimetuned$, $\epsilon_\text{min,BA}=\epstremorBArocketoverinceptiontimetuned$)
and the wavelet-feature MLP ($\epsilon_\text{min,mAP}=\epstremorAProcketovergp$, $\epsilon_\text{min,BA}=\epstremorBArocketovergp$).
InceptionTime with default hyperparameters is stochastically dominant over tuned InceptionTime ($\epsilon_\text{min,mAP} = \epstremorAPinceptiontimeoverinceptiontimetuned$, $\epsilon_\text{min,BA} = \epstremorBAinceptiontimeovergp$)
and the wavelet-based feature \gls{mlp} ($\epsilon_\text{min,mAP} = \epstremorAPinceptiontimeovergp$, $\epsilon_\text{min,BA} = \epstremorBAinceptiontimeovergp$).
InceptionTime with optimized hyperparameters is stochastically dominant over the wavelet \gls{mlp} ($\epsilon_\text{min,mAP} = \epstremorAPinceptiontimetunedovergp$, $\epsilon_\text{min,BA} = \epstremorBAinceptiontimetunedovergp$).
The confusion matrix in \cref{fig:confusionmatrix} highlights the class imbalance which results in all models predicting too low tremor severity. \Gls{rocket} tends to underestimate tremor more than InceptionTime but is also more accurate for zero or weak tremor.
{\sisetup{round-mode=figures,round-precision=3}%
\gls{rocket} displays the lowest \gls{mamae} (mean \num{0.605049047
}, S.D. \num{0.05978055}), followed by default InceptionTime (mean \num{0.76984629}, S.D. \num{0.075846933}
), tuned InceptionTime (mean \num{0.799789354}, S.D. \num{0.112201088}), and the wavelet \gls{mlp} (mean \num{0.94720854}, S.D. \num{0.047791374}).}
\begin{figure}
    \renewcommand{\rmfamily}{\sffamily}
    \centering
    \scalebox{0.9}{\import{figures/}{confusionmatrix.pgf}}
    \caption{Confusion matrix for the predictions of all classifiers based on GENEActiv smartwatch data. Out of the ten trained and evaluated classifiers, the fourth-best one according to mAP is selected, approximating the median performance.}
    \label{fig:confusionmatrix}
\end{figure}
\begin{figure}
    \centering
    \renewcommand{\rmfamily}{\sffamily}
    \resizebox{\textwidth}{!}{\import{figures}{relplot.pgf}}
    \caption{Reliability diagrams for all binary classifiers as a smoothed scatterplot of the groundtruth label $E\left[ y|f\right]$ vs. the predicted probability $f$.
    The red line represents the smoothed probabilities with thicker red lines representing a higher sample density.
    The shaded area represents the \SI{95}{\percent} confidence interval with darker shades representing a higher sample density.
    An upwards/downwards tick on the x-axis signifies positive/negative groundtruth labels for a predicted $f$.}
    \label{fig:reldiag}
\end{figure}

\subsubsection*{Bradykinesia}
\label{sec:resultsbradyga}
All classifiers substantially outperform random classifiers.
Default InceptionTime produces the best \gls{ap} scores for bradykinesia prediction (\cref{fig:genactivmodelcomp}).
However, the variability of InceptionTime scores (tuned and default) is much larger than the variability of scores produced by the other models.
InceptionTime with default hyperparameters is stochastically dominant over \gls{rocket} ($\epsilon_\text{min,mAP} = \epsbradykinesiaAPinceptiontimeoverrocket$, $\epsilon_\text{min,BA}=\epsbradykinesiaBAinceptiontimeoverrocket$)
and the wavelet \gls{mlp} ($\epsilon_\text{min,mAP} = \epsbradykinesiaAPinceptiontimeovergp$, $\epsilon_\text{min,BA}=\epsbradykinesiaBAinceptiontimeovergp$).
InceptionTime with tuned hyperparameters scores higher than \gls{rocket} ($\epsilon_\text{min,AP} = \epsbradykinesiaAPinceptiontimetunedoverrocket$, $\epsilon_\text{min,BA}=\epsbradykinesiaBAinceptiontimetunedoverrocket$)
and the wavelet \gls{mlp} ($\epsilon_\text{min,AP} = \epsbradykinesiaAPinceptiontimetunedovergp$, $\epsilon_\text{min,BA}=\epsbradykinesiaBAinceptiontimetunedovergp$).
\Gls{aso} provides no evidence for significant stochastic dominance of default InceptionTime over tuned InceptionTime ($\epsilon_\text{min,AP} = \epsbradykinesiaAPinceptiontimeoverinceptiontimetuned$, $\epsilon_\text{min,BA} = \epsbradykinesiaBAinceptiontimeoverinceptiontimetuned$).
A higher predicted probability tends to imply a higher chance of actual bradykinesia, as the calibration plots in~\cref{fig:reldiag} show.
ROCKET with softmax has an even distribution of scores while the other classifiers tend to predict either very high or very low scores (note the line thinning around $f=0$).

False positives are overrepresented during drawing for InceptionTime (458/168.7) and \gls{rocket} (550/150.9).
InceptionTime also has a higher-than-expected frequency of walking down a passage (279/178.8) and walking straight (131/93.2) among false positives.

\subsubsection*{Dyskinesia}
\label{sec:resultsdysk}
The three \gls{ap} scores barely exceed those expected from random classifiers (\cref{fig:genactivmodelcomp}).
\Gls{rocket} has a substantially higher \gls{ap} than the other classifiers.
The InceptionTime classifiers achieve higher \gls{ap} than the wavelet \gls{mlp}.
The InceptionTime models have the highest balanced accuracy, followed by \gls{rocket} and then the wavelet \gls{mlp}.
The default InceptionTime architecture shows high variability in terms of both \gls{ap} and balanced accuracy.
However, the ten samples yield a power of only \powerdyskinesiaAPinceptiontime{} for InceptionTime dyskinesia predictions and \powerdyskinesiaAPinceptiontimetuned{} for the tuned InceptionTime dyskinesia predictions, which is too low to enable comments on significance.
\Gls{rocket} is stochastically dominant over the wavelet \gls{mlp} ($\epsilon_\text{min,mAP} = \epsdyskinesiaAProcketovergp$, $\epsilon_\text{min,BA}=\epsbradykinesiaBArocketovergp$).
\Cref{fig:reldiag} shows that all classifiers predict very low dyskinesia probabilities, although \gls{rocket} again has the most homogenous prediction probability distribution.
Due to the low accuracy in dyskinesia detection, no further statements can be made about the quality of the calibration.

\subsection*{Misclassification Analysis}
{\sisetup{round-mode=figures,round-precision=3}
Misclassifications are analyzed for \gls{rocket} and InceptionTime with the default hyperparameters because InceptionTime with tuned and default hyperparameters show similar misclassification patterns.
The test dataset time series magnitude has a mean sample entropy of \num{1.0661607901436376} (S.D. \num{0.507125}, range \num{0.018731}--\num{2.245151}) and a mean standard deviation of \SI{1.218615}{\m\per\s\squared} (S.D. \SI{0.925058}{\m\per\s\squared}, range \num{0.060518}--\SI{4.285965}{\m\per\s\squared}).

{\sisetup{round-mode=figures,round-precision=3}
InceptionTime tremor false positives are strongly overrepresented (observed samples/expected samples) during nuts-and-bolts assembly (216/100) and drawing (200/135).
\Gls{rocket} false positives are strongly overrepresented during drinking (191/73.3), organizing papers (165/80.6), drawing (143/80.6), and nuts-and-bolts assembly (99/59.9).
Notably, for these tasks, the \gls{psd} in the \SIrange[round-mode=none]{4}{6}{\hertz} range is unaffected by the presence or absence of tremors.
In contrast, for the remaining tasks, patients with tremor show higher \gls{psd} in the \SIrange[round-mode=none]{4}{6}{\hertz} band than patients without tremor.
False-negative tremor estimation of InceptionTime occurs more frequently than expected when walking a narrow passage (135/70.04) and drawing (188/70.4).
False-negative tremor estimation of \gls{rocket} is more frequent than expected when walking a narrow passage (148/79.5) and walking straight (154/90.3).
Drawing has the highest complexity of all the tasks (mean sample entropy \rnum{1.864753}), while walking straight and down a passage has higher-than-average amounts of motion (mean standard deviations \SI{2.373923}{\m\per\s\squared} and \SI{1.708428}{\m\per\s\squared}).}

{\sisetup{round-mode=figures,round-precision=3}
When a patient has bradykinesia according to the groundtruth, the standard deviation of the motion time series magnitude is higher on average, indicating a greater degree of motion.
We find that true negative examples of bradykinesia time series have a higher average standard deviation (ROCKET \SI{1.429757}{\m\per\s\squared}, InceptionTime \SI{1.397880}{\m\per\s\squared}, tuned InceptionTime \SI{1.242009}{\m\per\s\squared}) than false positives (ROCKET \SI{0.809033}{\m\per\s\squared}, InceptionTime \SI{0.961606}{\m\per\s\squared}, tuned InceptionTime \SI{1.129852}{\m\per\s\squared}).
Similarly, true positives have a higher average standard deviation (ROCKET \SI{2.012657}{\m\per\s\squared}, InceptionTime \SI{1.958425}{\m\per\s\squared}) than false negatives (ROCKET \SI{1.520954}{\m\per\s\squared}, InceptionTime \SI{1.581119}{\m\per\s\squared}).}

As the performance of dyskinesia classifiers is only slightly better than random, dyskinesia misclassifications are expected to also be random, and no meaningful misclassification patterns can be identified.

\section*{Discussion}
\label{chap:discussion}
The results show that the time series classification approaches of InceptionTime and \gls{rocket} can learn to estimate \gls{pd} symptom severity from wrist accelerometer data during \gls{adl} with performance exceeding that of an \gls{mlp} with explicitly modeled wavelet features.

Dyskinesia is the most difficult \gls{pd} symptom to detect in ~\gls{adl} using wearable accelerometers and our studied approaches.
Our high accuracy is misleading due to class imbalance, and we only slightly outperform random classifiers.
In contrast, all approaches substantially outperform random classifiers for tremors and bradykinesia.
\Gls{rocket} substantially outperforms the other classifiers in terms of \gls{ap} for dyskinesia estimation
and has an acceptable margin over the random classifier baseline. \Gls{rocket}'s higher performance could be attributed to better performance of random kernels on smaller datasets, in comparison to learned kernels~\cite{dempster2020rocket}.
Furthermore, dyskinesia has a movement frequency slower than tremor and faster than bradykinesia, causing a higher overlap between dyskinesia and \gls{adl} in the frequency domain. This characteristic may have contributed to the lower performance observed as well. 
In comparison to the other symptoms, dyskinesia is highly dependent on context because it describes the patient moving when he/she does not want to move. Our dataset combines various \gls{adl}, which include high movement (typing) and low movement (sitting); accurate dyskinesia presence estimation would require ``detecting'' the current activity and assessing movement intensity relative to that activity.
Existing research has demonstrated better dyskinesia estimation, especially when adding gyroscopes~\cite{pfisterHighResolutionMotorState2020} and multiple sensor locations~\cite{hssayeniDyskinesiaEstimationActivities2021}, potentially even electromyography~\cite{cole2015wearablepd}.
Using multiple sensors can aid in distinguishing between symptomatic movements and \gls{adl}.
The sensors in the present work can only measure translational acceleration, but the twisting movements of dyskinesia are primarily rotational instead of translational\,---\,a rapidly rotating but relatively stationary wrist can occur in dyskinesia and will only lead to a small signal in the acceleration time series.
Thus, gyroscopes may be beneficial when the goal is to improve clinical performance.
Assuming that dyskinesia estimation is particularly challenging, deep learning models might require more training data to achieve reasonable performance, explaining the performance of InceptionTime lagging behind \gls{rocket}.

\Gls{rocket} and InceptionTime have similar performance for tremor estimation, and estimate tremor severity with an average error of less than one severity class (see \gls{mamae} results).
In contrast, the InceptionTime classifiers are substantially better at bradykinesia estimation.
InceptionTime tends to demonstrate the highest balanced accuracy, suggesting that InceptionTime might be slightly superior to \gls{rocket} when absolute predictions are more important than good calibration.
\Gls{rocket} and InceptionTime perform similarly on a variety of time series classification benchmarks~\cite{ruiz2020bakeoff,dempster2020rocket}, which fits the similar results for \gls{pd} motor symptom estimation.
InceptionTime demonstrates greater score variability across training runs than the other approaches, independent of whether automated hyperparameter tuning has been performed (see quantiles in \cref{fig:genactivmodelcomp}).
This variability aligns with the finding of the InceptionTime authors, who made it an ensemble of Inception networks due to the high variability of a single Inception network~\cite{fawaz2020inceptiontime}.
However, we show that for our problem, even ensembling cannot fully mitigate the variability of Inception networks for time series classification.
InceptionTime's sensitivity to random initialization may also contribute to the futility of automated hyperparameter optimization.
If InceptionTime reacts to several thousand trainable parameters that cannot be fully optimized by training, the impact of tuning additional hyperparameters may be minimal.
This is evidenced by the large variation of the model scores from the cross-validation.
\Gls{rocket}'s scores are remarkably stable, especially when compared with InceptionTime.

The misclassification patterns of InceptionTime and \gls{rocket} are similar.
Notably, false-positive tremor classifications occur more frequently than expected during tasks requiring fine motor coordination (drawing/writing, assembling nuts-and-bolts, organizing papers, and drinking).
Generally, tremor is characterized by motions between \SI{4}{\hertz} and \SI{6}{\hertz}\cite{colcherClinicalManifestationsParkinsons1999}. During fine-motor tasks, patients without tremor display more motion in this frequency band compared to those with tremor.
This suggests that tremors may hinder patients from performing the requested task, reducing the overall motion during task execution, and subsequently increasing the risk of false-positive tremor detection during fine motor tasks.
For bradykinesia, we find that misclassifications are associated with lower movement intensity.
Given that bradykinesia is the unintended slowness of movement, our classifiers may fail to detect bradykinesia (false negative) when the bradykinetic patient is moving less overall, because it may seem like intentionally slow movement.
Conversely, the models may misclassify intentionally slow movement as bradykinesia (false positive).

Models for all symptom severities demonstrate improved performance with longer window lengths up to \SI{30}{\s} (see \cref{fig:hyperscatter}).
In our case, longer time series may give the classifiers a better chance of separating the superimposed \gls{pd} symptoms from the \gls{adl}.
Using random slices from a time series with a window length of \SI{90}{\percent} of the original time-series length has been shown to improve deep learning time series classification~\cite{leguennec2016augmentation}, but we found no research comparing this to windows shorter than \SI{90}{\percent}.
In the context of \gls{pd} deep learning, some researchers have used \SI{30}{\second} windows for tremor and dyskinesia~\cite{cole2015wearablepd}, and others have used \SI{5}{\s} windows for bradykinesia~\cite{eskofierRecentMachineLearning2016}. None of those studies provide results for experiments on window length.

Extensive hyperparameter tuning via random search does not improve performance compared to default InceptionTime hyperparameters on a held-out test dataset.
Vastly different architectures perform similarly.
For instance, the tuned InceptionTime tremor classifier has \num{7499} trainable parameters, while the default four-class InceptionTime has \num{490564} trainable parameters, yet both achieve similar scores.
This similarity in test performance pre- and post-tuning may suggest overfitting of hyperparameters to the cross-validation sets.
Alternatively, hyperparameters---with the exception of window length---may not substantially impact the InceptionTime performance.

When comparing with the established literature, InceptionTime, \gls{rocket}, and the wavelet \gls{mlp} score lower as per \cref{tab:litcomp}, but there are very few works with five classes and \gls{adl}, limiting the possibility of comparison.
Using engineered features and a random forest~\cite{loniniWearableSensorsParkinsons2018} yields higher \gls{auroc} than our approaches and wavelets with Gaussian processes~\cite{langMultiLayerGaussianProcess2019} are more accurate.
Other works either had patients at rest~\cite{wagnerWaveletbasedApproachMonitoring2017} or performing activities that were optimized for tremor detection~\cite{polvorinos-fernandezEvaluationPerformanceWearables2024,sigchaAutomaticRestingTremor2021,legaria-santiagoComputerModelsEvaluating2022} instead of the \gls{adl} in our research.
Further work used multiple or different sensor locations~\cite{papadopoulosDetectingParkinsonianTremor2019,papadopoulosLeveragingUnlabelledData2023} and/or had different numbers of classes~\cite{reardonWearableSensorConfigurations2024b,eversPassiveMonitoringParkinson2025
,wuMultiInstanceLearningParkinsons2025}.
\begin{table}
    \caption{Comparison of our mean results (highlighted) with values from the literature that use similar motions and at least the same number of classes.}
    \label{tab:litcomp}
    \newcommand{\ours}{\color{OliveGreen}}
    \sisetup{multi-part-units=single,separate-uncertainty,table-format=2.1(3)}
    \resizebox{\textwidth}{!}{\centering\begin{tabular}{llrSSSS}
\toprule
 & & &  \multicolumn{4}{c}{Metric $\pm$ Std. Dev. / \%} \\
\cmidrule(lr){4-7}
\multicolumn{1}{@{}l}{Method} & {Motions} & {Classes} & {Accuracy} & {BA} & {AUROC} & {F1} \\
\midrule
\multicolumn{1}{@{}l}{\itshape{Tremor}} \\
hand accel.+gyro; features+RF\cite{loniniWearableSensorsParkinsons2018} & free & 5\hphantom{*} &  &  & \bfseries 77.6 &  \\
watch accel.+gyro; wavelet+GP\cite{langMultiLayerGaussianProcess2019} & free & 5\hphantom{*} & \bfseries 82.6 &  &  &  \\
\ours
watch accel.; wavelet MLP & 18 prescribed ADL & 4\hphantom{*} & 59.0\pm1.5 & 41.1\pm1.6 & 66.6\pm1.2 & 38.1\pm1.5 \\
\ours
watch accel.; tuned InceptionTime & 18 prescribed ADL & 4\hphantom{*} & 64.8\pm3.9 & 41.3\pm4.6 & 71.7\pm2.9 & 37.6\pm3.4 \\
\ours
watch accel.; ROCKET & 18 prescribed ADL & 4\hphantom{*} & 71.0\pm0.6 & 44.6\pm1.5 & 73.0\pm0.6 & 40.5\pm1.0 \\
\ours
watch accel.; default InceptionTime & 18 prescribed ADL & 4\hphantom{*} & 65.4\pm2.4 & \bfseries 46.1\pm2.5 & 72.8\pm2.1 & \bfseries 41.2\pm2.2 \\
\midrule
\multicolumn{1}{@{}l}{\itshape{Bradykinesia}} \\
hand accel.+gyro; features+RF\cite{loniniWearableSensorsParkinsons2018} & free & 5\hphantom{*} &  &  & 69.0 &  \\
watch accel.+gyro; wavelets+GP\cite{langMultiLayerGaussianProcess2019} & free & 5\hphantom{*} & 61.2 &  &  &  \\
watch accel.; features+SVM\cite{habetsRapidDynamicNaturalistic2021} & free & 2\hphantom{*} & 66.9\pm10.0 &  & 62.4\pm9.0 &  \\
watch accel.+gyro; wavelet+GP\cite{endo2018gp} & free & 5\hphantom{*} & 69.3 &  &  &  \\
watch accel.; wavelet+SVM\cite{wagnerWaveletbasedApproachMonitoring2017} & 20 prescibed ADL & 2\hphantom{*} & 82.1 & 67.1 & 70.0 & \bfseries 88.9 \\
\ours
watch accel.; wavelet MLP & 18 prescribed ADL & 2\hphantom{*} & 71.3\pm1.8 & 69.2\pm1.6 & 77.1\pm1.6 & 55.5\pm2.0 \\
watch accel.; features+LR\cite{eversRealLifeGaitPerformance2020} & free & 2\hphantom{*} & 76.0 & 72.0 &  &  \\
\ours
watch accel.; ROCKET & 18 prescribed ADL & 2\hphantom{*} & 74.7\pm0.9 & 72.5\pm1.1 & 79.9\pm0.8 & 59.8\pm1.4 \\
\ours
watch accel.; tuned InceptionTime & 18 prescribed ADL & 2\hphantom{*} & 77.5\pm1.2 & 75.0\pm2.6 & 83.6\pm1.8 & 62.8\pm3.1 \\
\ours
watch accel.; default InceptionTime & 18 prescribed ADL & 2\hphantom{*} & 75.2\pm1.4 & 75.5\pm0.9 & \bfseries 84.1\pm1.2 & 63.1\pm1.2 \\
 watch accel.+gyro; timeseries-to-image+CNN\cite{pfisterHighResolutionMotorState2020} & free & 3* & \bfseries 82.6 & \bfseries 76.8 &  & 66.5 \\
\midrule
\multicolumn{1}{@{}l}{\itshape{Dyskinesia}} \\
watch accel.+gyro; wavelets+GP\cite{langMultiLayerGaussianProcess2019} & free & 5\hphantom{*} & 49.2 &  &  &  \\
watch accel.+gyro; wavelet+GP\cite{endo2018gp} & free & 5\hphantom{*} & 59.0 &  &  &  \\
\ours
watch accel.; wavelet MLP & 18 prescribed ADL & 2\hphantom{*} & 88.4\pm0.7 & 51.5\pm2.1 & 56.3\pm3.3 & 9.2\pm4.3 \\
\ours
watch accel.; ROCKET & 18 prescribed ADL & 2\hphantom{*} & \bfseries 93.2\pm0.2 & 53.1\pm0.9 & \bfseries 70.9\pm1.5 & 11.7\pm3.1 \\
\ours
watch accel.; default InceptionTime & 18 prescribed ADL & 2\hphantom{*} & 91.1\pm1.0 & 55.7\pm3.7 & 64.4\pm7.7 & 18.2\pm8.3 \\
\ours
watch accel.; tuned InceptionTime & 18 prescribed ADL & 2\hphantom{*} & 90.8\pm1.8 & 55.7\pm4.8 & 62.5\pm3.7 & 18.5\pm11.0 \\
watch accel.+gyro; timeseries-to-image+CNN\cite{pfisterHighResolutionMotorState2020} & free & 3* & 81.6 & \bfseries 77.0 &  & \bfseries 69.0 \\
\midrule
\end{tabular}
}
    \small *A single model outputs either no PD, bradykinesia, or dyskinesia
\end{table}
With regards to bradykinesia detection, two out of seven related studies report better accuracy, one out of three reports better balanced accuracy, none report better \gls{auroc}, and all report better F1.
Our approaches achieve higher \gls{ba} and \gls{auroc} but lower accuracy and F1 than a study that also looked at many \glspl{adl} on a dataset very similar to the \gls{mjff} dataset~\cite{wagnerWaveletbasedApproachMonitoring2017}.
We also report higher accuracy than using explicitly modeled features when patients are free to move around~\cite{endo2018gp,langMultiLayerGaussianProcess2019,loniniWearableSensorsParkinsons2018}--but these studies also have more classes.
An interesting approach of converting acceleration time series into images for classification by a vision \gls{cnn} demonstrated slightly better scores~\cite{pfisterHighResolutionMotorState2020}.
However, they had one classifier that categorized motions into bradykinetic, healthy, or dyskinetic, in contrast to our separate binary classifiers for bradykinesia and dyskinesia.
Many other related works for bradykinesia detection cannot be compared directly because they have restricted patients motions~\cite{eskofierRecentMachineLearning2016,sigchaBradykinesiaDetectionParkinsons2022} or used other sensors~\cite{parkEvaluationParkinsonianBradykinesia2021,pastorino2011bradymultiparam,pulliamContinuousAssessmentLevodopa2018}.
In accordance with our low performance and poor calibration for dyskinesia detection, all related works (see \cref{tab:litcomp}) report higher balanced accuracy and F1.
Our high accuracy comes from the tendency to predict no dyskinesia and the skewed class distribution (see \cref{fig:confusionmatrix}).
However, in absolute terms, all studies except one~\cite{pfisterHighResolutionMotorState2020} report low accuracies (under \SI{60}{\percent})\cite{endo2018gp,langMultiLayerGaussianProcess2019}.
Several other related works are excluded from the numeric comparison due to differing sensors~\cite{hssayeniDyskinesiaEstimationActivities2021,pulliamContinuousAssessmentLevodopa2018} and activities~\cite{wagnerWaveletbasedApproachMonitoring2017}.
We note that comparisons with related work on detecting \gls{pd} from wearable accelerometers are challenging because not all studies report all relevant metrics. Furthermore, datasets and sensors differ, and the classification task may be framed differently.
Finally, many scores reported in existing literature lack an uncertainty estimate, and there are subtleties in how multiclass metrics such as \gls{auroc} are calculated.

The \gls{mjff} dataset gives a single symptom severity label even if symptoms fluctuate during the task.
Some works address this problem of having whole time series labeled with a single symptom despite symptom fluctuations by using so-called \emph{multiple-instance} learning~\cite{papadopoulosDetectingParkinsonianTremor2019,papadopoulosLeveragingUnlabelledData2023} and/or manually examining the time series signals to modify the class labels~\cite{papadopoulosLeveragingUnlabelledData2023}.
Generally, machine learning applied to manually generated features is able to outperform InceptionTime and \gls{rocket}, especially when the patient motions are constrained.
All high-performing related work with raw data either used semi-supervised learning~\cite{papadopoulosLeveragingUnlabelledData2023} or data augmentation with domain-specific knowledge~\cite{pfisterHighResolutionMotorState2020,um2017pdaugmentation}.
Specially designed time series ordinal regression approaches outperform nominal classifiers~\cite{guijorubio2020ordvsnomtsc} and may be worth investigating to predict \gls{mdsupdrs} scores.
Treating symptom severity prediction as a non-linear regression problem models the clinical reality most closely, however.

We avoid selection bias in the classifier scoring by evaluating a held-out test dataset while developing models on a validation dataset with cross-validation.
Repeating the training of each final model ten times ensures the reliability of the results and provides insights into the score distribution.
The significance tests in our work provide statistically rigorous analysis and reduce bias, albeit with some caveats.
Significance testing with \gls{aso} works better when more different sources of variation are considered~\cite{ulmer2022deepsignificance}.
Although we vary the random initialization, we do not re-shuffle the training data, sub-sample it, or modify the train-validate-test split.
In other words, when considering the data split as nested $k$-fold grouped stratified cross-validation, the outer loop has a count of only one.
Hence, some findings reported as significant may not generalize to applications of similar pipelines~\cite{bouthillier2021variance}.
Our results hinge on the Levodopa Response Study dataset, which encompasses 28 patients, but strong tremors are exceedingly rare.
Accordingly, when grouping by patients, it is nearly impossible to create training, validation, and test datasets containing tremor severity four while retaining representative distributions for other symptoms and severities across the datasets (see \cref{tab:datasplit}).
Hence, classifiers have very few examples from which to learn about strong tremors.
Furthermore, the cross-validation-based automated hyperparameter search cannot consider scores for the strongest tremors as only one patient across the training and validation data has these strong tremors.
Our use of data stratification minimizes the impact of the class imbalance on the reported results.
The limited dataset size also means that the evaluation of performance cannot be separated by task while preserving statistical validity.
Furthermore, the Levodopa Response Study's use of different raters for different tasks and patients may cause systematic errors or bias in the labels.
However, incorrect labels would affect all classifiers, and our conclusions regarding substantial relative performance differences remain valid.
Finally, bias might arise in future real-world applications because wearing the smartwatch on the predominantly affected right hand may conflict with cultural and individual preferences for wearing watches on the left (or non-dominant) hand.

Our hypothesis was that the state-of-the-art end-to-end time-series classification approaches would also perform well for estimating \gls{pd} symptom severity estimation during \gls{adl}.
This hypothesis is supported by comparisons with our wavelet-based baseline, which demonstrate the ability of both InceptionTime and \gls{rocket} to implicitly learn discriminative features from raw tri-axial wrist accelerometry.

\section*{Conclusion}
\label{sec:conclusion}
We compared InceptionTime, \gls{rocket}, and an \gls{mlp} operating on wavelet-based features for predicting \gls{pd} symptom severity from wrist-worn accelerometer data.
InceptionTime is comparatively better suited to predicting tremor and bradykinesia, significantly outperforming the simpler wavelet-based MLP approach and slightly surpassing ROCKET overall.
However, the performance of InceptionTime varies greatly depending on the random initialization of weights.
\Gls{rocket} represents the most effective estimator for dyskinesia and is the only model that substantially outperforms random classifiers, although dyskinesia remains the most challenging symptom to detect.
Our extensive hyperparameter tuning involved the evaluation of 900 model configurations, and the results indicate that InceptionTime is unlikely to benefit substantially from architectural modifications.
Despite InceptionTime and \gls{rocket} representing the state of the art in time series classification with automated feature extraction, certain customized machine learning approaches, especially those employing manual feature engineering, still achieve superior performance for \gls{pd} symptom severity prediction. To the best of our knowledge, this study constitutes the first application of \gls{rocket} and InceptionTime to the prediction of PD motor symptoms from wrist accelerometry data.

\section*{Acknowledgements}
We thank the Michael J. Fox Foundation for funding the MJFF Levodopa Response Study and providing the dataset used for this paper.
This work
was partially funded by the European Research Council (ERC) Consolidator Grant ``Safe data-driven control for human-centric systems (CO-MAN)" under grant agreement number 864686,
by the Federal Ministry of Education and Research of Germany (BMBF) in the program of “Souverän. Digital. Vernetzt.” (
Joint project 6G-life, project identification number: 16KISK002),
and by the Federal Ministry of Education and Research of Germany and the Free State of Bavaria under the Excellence Strategy of the Federal Government and the States (TUM Innovation Network eXprt).

\section*{Author contributions statement}
C.D. and N.D. conceived and conducted the experiments. S.E., N.D., and S.H. developed the research idea. C.D. drafted the manuscript and drew the figures. All authors contributed to and reviewed the manuscript.

\section*{Data availability statement}
The data that support the findings of this study are available from the \gls{mjff} under \href{https://doi.org/10.7303/syn20681023}{https://doi.org/10.7303/syn20681023}.

\section*{Additional information}
The authors declare no competing interests.
Our source code is available from \href{https://github.com/cedricdonie/tsc-for-wrist-motion-pd-detection}{https://github.com/cedricdonie/tsc-for-wrist-motion-pd-detection}.

\end{document}